\documentclass[10pt,journal,compsoc]{IEEEtran}

\ifCLASSOPTIONcompsoc
\usepackage[nocompress]{cite}
\else
\usepackage{cite}
\fi

\ifCLASSINFOpdf
\usepackage[pdftex]{graphicx}
\usepackage{multirow}
\usepackage{amsmath}
\usepackage{amssymb}
\usepackage{booktabs}
\usepackage{caption}

\usepackage{xcolor} 
\usepackage{color} 
\usepackage{verbatim} 
\usepackage{diagbox}
\usepackage[letterpaper=true,colorlinks,citecolor=blue,linkcolor=blue]{hyperref}
\else
\fi
\usepackage{graphicx}  
\usepackage{float}
\usepackage{subfigure} 

\usepackage{bm}
\usepackage{gensymb}
\usepackage{xspace}

\newcommand{\ie}{i.e.\xspace}

\begin{document}

\title{DreamCraft3D++: Efficient Hierarchical 3D Generation with Multi-Plane Reconstruction Model}

\author{Jingxiang Sun,
        Cheng Peng,
        Ruizhi Shao,
        Yuan-Chen Guo,
        Xiaochen Zhao,
        Yangguang Li,
        YanPei Cao,
        Bo Zhang,
        Yebin Liu,~\IEEEmembership{Member,~IEEE}
\\{\ttfamily \url{https://dreamcraft3dplus.github.io/}}
\IEEEcompsocitemizethanks{\IEEEcompsocthanksitem Jingxiang Sun, Cheng Peng, Ruizhi Shao, Xiaochen Zhao and Yebin Liu are with Department of Automation, Tsinghua University, Beijing, China.
E-mail: {\{starkjxsun, chengpeng002, jia1saurus\}@gmail.com; zhaoxc19@mails.tsinghua.edu.cn; liuyebin@mail.tsinghua.edu.cn}
\IEEEcompsocthanksitem Yuan-chen Guo, Yangguang Li and Yanpei Cao are with VAST, Beijing, China.
E-mail: {\{imbennyguo, liyangguang256, caoyanpei\}@gmail.com}
\IEEEcompsocthanksitem Bo Zhang is with Zhejiang University, Hangzhou, China.
E-mail: {bo.zhang@zju.edu.cn}
\IEEEcompsocthanksitem Corresponding author: Yebin Liu.
}
}



\newcommand{\mname}{ACLRNet}
\markboth{submit to IEEE Transactions on Pattern Analysis and Machine Intelligence,~Vol.~XX, No.~XX, XX~2024}%
{Shell \MakeLowercase{\textit{Zamir et al.}}: Bare Demo of IEEEtran.cls for Computer Society Journals}

\IEEEtitleabstractindextext{
\begin{abstract}
We introduce DreamCraft3D++, an extension of DreamCraft3D that enables efficient high-quality generation of complex 3D assets. DreamCraft3D++ inherits the multi-stage generation process of DreamCraft3D, but replaces the time-consuming geometry sculpting optimization with a feed-forward multi-plane based reconstruction model, speeding up the process by 1000x. For texture refinement, we propose a training-free IP-Adapter module that is conditioned on the enhanced multi-view images to enhance texture and geometry consistency, providing a 4x faster alternative to DreamCraft3D's DreamBooth fine-tuning. Experiments on diverse datasets demonstrate DreamCraft3D++'s ability to generate creative 3D assets with intricate geometry and realistic 360° textures, outperforming state-of-the-art image-to-3D methods in quality and speed. The full implementation will be open-sourced to enable new possibilities in 3D content creation.
\end{abstract}

}

\maketitle
\IEEEdisplaynontitleabstractindextext

\IEEEpeerreviewmaketitle


\section{Introduction}\label{sec:introduction}

\begin{figure*}[tp]
\begin{center}
\includegraphics[width=0.9\linewidth]{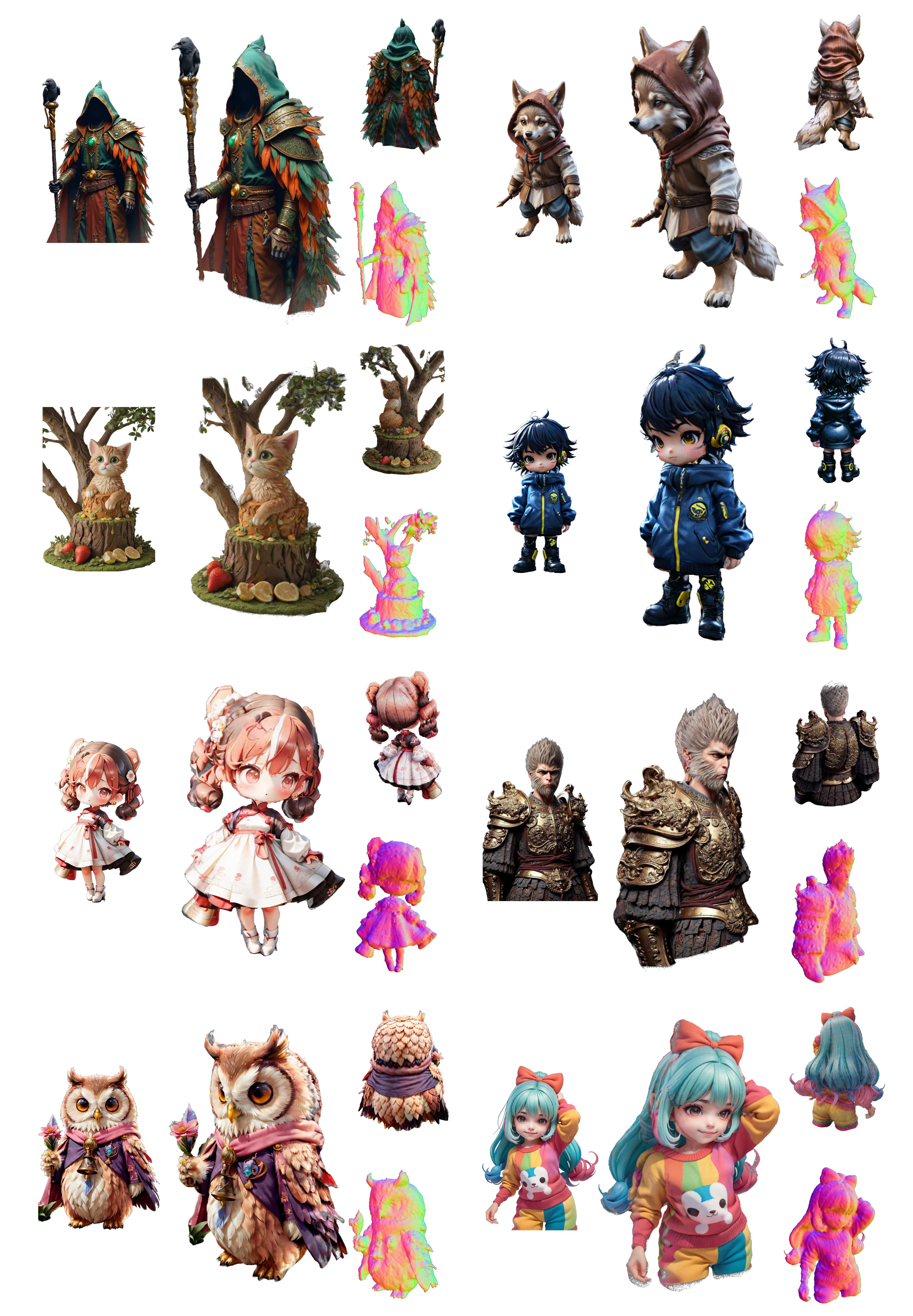}
\caption{By lifting 2D images to 3D, DreamCraft3D++ achieves 3D generation with rich details and
holistic 3D consistency. Please refer to the demo video for more results.}
\label{fig:teaser}
\end{center}
\end{figure*}

\IEEEPARstart{T}{he} remarkable success of 2D generative modeling~\cite{saharia2022photorealistic,ramesh2022hierarchical,rombach2022high,gu2022vector} has profoundly shaped the way that we create visual content. 3D content creation, which is crucial for applications like games, movies and virtual reality, still presents a significant challenge for deep generative networks. While 3D generative modeling has shown compelling results for certain categories~\cite{wang2023rodin,chan2022efficient,zhang2023avatarverse}, generating general 3D objects remains formidable due to the lack of extensive 3D data. 

Recent research in 3D generation has evolved into two main paradigms: 2D lifting~\cite{poole2022dreamfusion,lin2023magic3d,wang2023prolificdreamer,qian2023magic123,sun2023dreamcraft3d} and feed-forward 3D generation~\cite{zhang2024clay,wu2024direct3d,hong2023lrm,PF-LRM,Tang2024LGMLM,Xu2023DMV3DDM}. 2D lifting methods leverage pretrained 2D vision models to guide 3D optimization, with DreamFusion~\cite{poole2022dreamfusion} introducing Score Distillation Sampling (SDS) loss to align 3D renderings with text-conditioned image distributions. Subsequent works have improved photo-realism through stage-wise optimization and enhanced distillation losses. However, these approaches often struggle with complex content synthesis and suffer from the "Janus issue," where individual renderings appear plausible but lack holistic consistency. While recent advancements have significantly improved quality and speed, SDS-based methods remain computationally expensive, typically requiring minutes to hours for single object generation.

Recent advancements in the feed-forward 3D generation~\cite{zhang2024clay,wu2024direct3d,hong2023lrm,PF-LRM,Tang2024LGMLM,Xu2023DMV3DDM} have significantly improved efficiency through the use of 3D large reconstruction models(LRM)~\cite{hong2023lrm}. These approaches typically employ a two-step process: first, generating multi-view images from a single input image or text prompt, and then directly regressing 3D representations from these generated images using sparse-view reconstructors. However, these methods invariably require 3D shapes or multi-view data for training, which presents challenges when generating in-the-wild 3D assets due to the relative scarcity of diverse 3D data compared to 2D data. Furthermore, the reconstruction results often exhibit limitations, such as a lack of fine geometric structures and detailed texture patterns.

To recap, DreamCraft3D~\cite{sun2023dreamcraft3d} falls within the category of 2D lifting and draw inspiration from the manual artistic process, breaking down the challenging 3D generation into manageable steps. Starting with a high-quality 2D reference image generated from a text prompt, DreamCraft3D lifts it into 3D via stages of geometry sculpting and texture boosting. In the geometry sculpting stage, DreamCraft3D adopts joint 2D-3D SDS loss for novel views and photometric loss at the reference view for producing plausible and consistent 3D geometry. In the texture boosting stage, in order to obtain consistent texture, a pretrained text-to-image diffusion model is finetuned on multi-view renderings of the 3D instance, resulting a personalized 3D-aware generative prior for texture boosting. 

While DreamCraft3D significantly improves 3D generation quality, its main drawback is the lengthy processing time, requiring approximately 3 hours per case. In this paper, we introduce DreamCraft3D++, an enhanced extension of DreamCraft3D that enables efficient and high-quality complex 3D asset generation. Our approach maintains the high generation quality of DreamCraft3D while accelerating the process by a factor of 20, reducing the generation time from three hours to just 10 minutes. Through a detailed analysis of the optimization time for individual stages, we identified that the geometry sculpting stage accounts for over 70\% of the total time, as optimizing 3D structures from scratch without any shape prior is computationally expensive. Inspired by \cite{hong2023lrm}, we propose training a large reconstruction model to provide a coarse textured mesh, thereby replacing the original time-consuming geometry sculpting stage. \IEEEpubidadjcol

To achieve this, our approach utilizes multi-view images as input to a UNet-based architecture for predicting pixel-aligned multi-plane features, which is crucial for bridging 2D and 3D. Our method diverges from current counterparts~\cite{PF-LRM, hong2023lrm, xu2024instantmesh} in two key aspects. Firstly, our approach predicts a nonorthogonal multi-plane representation from fixed-view images, instead of predicting orthogonal axis-aligned planes (e.g. tri-planes) in \cite{ hong2023lrm, PF-LRM, mueller2022instant,zou2023triplane,Li2023Instant3DFT}. We opt for nonorthogonal planes instead of orthogonal ones because directly learning generalized canonical 3D features (e.g., axis-aligned triplanes or grids) from 2D posed images is difficult, particularly given limited training data and network capacity. By predicting nonorthogonal image-aligned plane features, the network can concentrate more effectively on translating 2D high-frequency details into 3D representations. Secondly, we leverage a U-Net convolutional architecture to map input images to multi-planes, exploiting the strong pixel-level alignment between input and output. The substantial bandwidth capacity of our U-Net enables direct transformation of multi-view images into multi-planes, yielding highly detailed results. Additionally, we incorporate normal maps into the reconstruction network, enhancing the model's comprehension of spatial relationships and geometry.




 From the generated multi-plane features, we further decode Flexicubes~\cite{shen2023flexible} parameters and predict the textured mesh. In contrast to NeRF's memory-intensive volume rendering, our approach leverages efficient mesh rasterization, enabling the use of full-resolution images and additional geometric information for supervision. This results in superior reconstruction quality. While alternative 3D representations such as Gaussian Splatting~\cite{kerbl2023gaussian} also facilitate efficient high-resolution rendering, they lack explicitly defined surfaces, making them less suitable for geometric modeling.


The output textured meshes of the above geometry sculpting stage are highly consistent to the multi-view input images, though, still suffer from detail distortion because 1) limited network generalbility due to lacking data; 2) inconsistent and blurry generated multi-view renderings from mv diffusion models. Therefore, we introduce a novel refinement algorithm to enhance both texture and geometry concurrently. The core insight lies in that the pretrained 2D diffusion priors lack object awareness and lead to inconsistent optimization, shared insight with DreamCraft3D. But different from it which finetunes a pretrained diffusion model with the multi-view image renderings using DreamBooth, we instead integerate a lightweight image prompt
adaptation module, named IP-Adapter\cite{ye2023ip}, enabling the model to form a concept about the scene’s surrounding views. Different from DreamBooth, Ip-adapter is training-free so it is much more efficient. Meanwhile, conditioning the IP-adapter solely on the source image can lead to ambiguity and the texture multi-head problem. To address this, we condition the IP-adapter on both the source image and augmented multi-view renderings. During training, the appropriate IP-adapter image embedding is dynamically selected based on the position of the randomly sampled camera.

As shown in Figure~\ref{fig:teaser}, our method is capable of producing creative 3D assets with intricate geometric structures and realistic textures rendered coherently in 360\degree. Compared to optimization-based approaches ~\cite{poole2022dreamfusion,lin2023magic3d}, our method offers substantially improved texture and efficiency. Meanwhile, compared to image-to-3D  techniques~\cite{tang2023make,qian2023magic123}, our work excels at producing unprecedentedly realistic renderings in 360\degree renderings.
These results suggest the strong potential of DreamCraft3D++ in enabling new creative possibilities in 3D content creation. 
The full implementation will be made publicly available. 


While preserving the core multi-stage framework and object-awareness, this paper extends DreamCraft3D, the conference version of this work in the following key aspects: 
\begin{itemize}
    \item For the coarse geometry sculpting stage, we introduce a feed-forward multi-plane based large reconstruction model to replace time-expensive optimization in DreamCraft3D, speeding up 1000 times with comparable results;
    \item We introduce a training-free IP-Adapter to enhance texture and geometry, achieving comparable results to DreamCraft3D's DreamBooth fine-tuning while being 4 times faster. Our IP-Adapter's dynamic embedding selection based on camera position addresses texture inconsistency and preserves fidelity, offering an efficient alternative to DreamCraft3D's approach.
    \item Compared to DreamCraft3D, we conduct experiments on a wider range of datasets, demonstrating the robustness and superiority of our model over other image-to-3D methods.
\end{itemize}

\section{Related works}

In this work, we focus on the task of generating high-quality geometric and richly textured 3D models from single images. We categorize the main research efforts in this area into three aspects: Novel-View Synthesis, which involves generating views from different perspectives of an object based on input text or generating new viewpoint views from a single input image; Progressively Optimized Reconstruction, which leverages 2D generative models to progressively optimize implicit 3D neural fields; and Feed-Forward Generation, which involves generating 3D representations in a feed-forward manner based on given text or image instructions.

\subsection{Novel-View Synthesis}

Recently, direct novel views synthesis(NVS) from single images of a 3D object has been explored, these works~\cite{sargent2023vq3d,skorokhodov20233d,xiang20233d} often rely on a pretrained monocular depth prediction model to synthesize view-consistent images. While some models achieve photo-realistic renderings for ImageNet categories, they struggle with large views. Recent attempts ~\cite{watson2022novel,liu2023zero,shi2023zero123++,Voleti2024SV3DNM,Shi2023MVDreamMD,Wang2023ImageDreamIM,one-2-3-45,qiu2024richdreamer} training view-dependent diffusion models on 3D data show promising results for open-domain novel view synthesis but, as inherently 2D models, can't ensure perfect view consistency.

Based on this rationale, diffusion models for NVS are utilized as inputs for feed-forward 3D generation models, aiming to leverage the generative capabilities of large models to eliminate inconsistencies present in multi-view images while preserving reasonable texture and geometric information found in those views. Inspired by these works, we adopt the generative outputs of the multi-view generation model Zero123++~\cite{shi2023zero123++} as prior inputs for our mp-lrm model.

\subsection{Progressively Optimized Reconstruction}

Progressively Optimized Reconstruction improve a 3D scene representation by seeking guidance using established 2D text-image foundation models. Early works~\cite{jain2021dreamfields,lee2022understanding,hong2022avatarclip} utilize the pretrained CLIP~\cite{radford2021learning} model to maximize the similarity between rendered images and text prompt. DreamFusion~\cite{poole2022dreamfusion} and SJC~\cite{sjc}, on the other hand, propose to distill the score of image distribution from a pretrained diffusion model and demonstrate promising results. Recent works have sought to further enhance the texture realism via coarse-to-fine optimization~\cite{lin2023magic3d, chen2023fantasia3d}, improved distillation loss~\cite{wang2023prolificdreamer,liu2023unidream,huang2023epidiff}, shape guidance~\cite{metzer2023latent} or lifting NVS 2D images to 3D~\cite{deng2023nerdi,tang2023make,qian2023magic123,liu2023one,huang2023customize,Liu2023SyncDreamerGM,Voleti2024SV3DNM,Tang2023DreamGaussianGG,sun2023dreamcraft3d}. 
Recently,~\cite{raj2023dreambooth3d} proposes to finetune a personalized diffusion model for 3D consistent generation. However, producing globally consistent 3D remains challenging. DreamCraft3D~\cite{sun2023dreamcraft3d} meticulously design 3D priors through the whole hierarchical generation process, achieving unprecedented coherent 3D generation. However, 
the process necessitates approximately three hours per case, which is highly inefficient.

\subsection{Feed-forward Generation}

3D generative models have been intensively studied to generate 3D assets without tedious manual creation. 
Generative adversarial networks (GANs)~\cite{chan2021pi, chan2022efficient, chan2021pi, xie2021style, zeng2022lion, skorokhodov20233d, gao2022get3d, tang2022explicitly, xie2021style, Sun_2023_CVPR, ide3d} have long been the prominent techniques in the field.  
Auto-regressive models have been explored~\cite{sanghi2022clip,mittal2022autosdf,yan2022shapeformer,zhang20223dilg,yu2023pushing},  which learn the distribution of these 3D shapes conditioned on images or texts. 
Diffusion models~\cite{wang2023rodin,cheng2023sdfusion,li2023diffusion,nam20223d,zhang20233dshape2vecset,nichol2022point,jun2023shap,bautista2022gaudi,gupta20233dgen,long2023wonder3d,liu2023pi3d,Zhang2024CLAYAC,Wu2024Direct3DSI} have also shown significant recent success in learning probabilistic mappings from text or images to 3D shape latent. 

Recently, a series of large-scale 3D reconstruction models based on transformers have emerged~\cite{hong2023lrm,Li2023Instant3DFT,PF-LRM, Xu2023DMV3DDM,zou2023triplane,gslrm2024,Siddiqui2024Meta3A}. Some of these models use single images~\cite{hong2023lrm,PF-LRM,Xu2023DMV3DDM,zou2023triplane,gslrm2024}, while others utilize multiple images generated by NVS models~\cite{Li2023Instant3DFT,Siddiqui2024Meta3A}. These models employ an end-to-end approach to generate implicit 3D representations using triplane NeRF. In addition, UNet has also been proven effective in generating multiplane and Gaussian representations~\cite{Wang2024CRMSI,Tang2024LGMLM}.

However, all these methods require 3D shapes or multi-view data for training, raising challenges when generating in-the-wild 3D assets due to the scarcity of diverse 3D data~\cite{chang2015shapenet,deitke2023objaverse,wu2023omniobject3d} compared to 2D. At the same time, the reconstruction results of these methods often face issues such as the lack of fine geometric structures and the lack of exquisite texture patterns.









\section{Preliminaries}
DreamFusion~\cite{poole2022dreamfusion} achieves text-to-3D generation by utilizing a pretrained text-to-image diffusion model $\bm{\epsilon}_\phi$ as an image prior to optimizing the 3D representation parameterized by $\theta$. 

The image $\bm{x}=g(\theta)$, rendered at random viewpoints by a volumetric renderer, is expected to represent a sample drawn from the text-conditioned image distribution $p(\bm{x}|y)$ modeled by a pretrained diffusion model. The diffusion model $\phi$ is trained to predict the sampled noise $\bm{\epsilon}_\phi(\bm{x}_t;y,t)$ of the noisy image $\bm{x}_t$ at the noise level $t$, conditioned on the text prompt $y$. A \emph{score distillation sampling} (SDS) loss encourages the rendered images to match the distribution modeled by the diffusion model. Specifically, the SDS loss computes the gradient:
\begin{equation}
\nabla_{\theta}\mathcal{L}_\textnormal{SDS}(\phi, g(\theta))=\mathbb{E}_{t, \bm{\epsilon}}\Big[\omega(t)(\bm{\epsilon}_{\phi}(\bm{x}_{t};y,t)-\bm{\epsilon})\frac{\partial \bm{x}}{\partial \theta}\Big],
\label{eq:sds}
\end{equation}
which is the per-pixel difference between the predicted and the added noise upon the rendered image, where $\omega(t)$ is the weighting function.

One way to improve the generation quality of a conditional diffusion model is to use the classifier-free guidance (CFG) technique to steer the sampling slightly away from the unconditional sampling, \ie,  $\epsilon_\phi(\bm{x}_t;y,t) + \omega\epsilon_\phi(\bm{x}_t;y,t) - \omega\epsilon_\phi(\bm{x}_t,t,\varnothing)$, where $\varnothing$ represents the ``empty'' text prompt. Typically, the SDS loss requires a large CFG guidance weight for high-quality text-to-3D generation, yet this will bring side effects like over-saturation and over-smoothing~\cite{poole2022dreamfusion}. 

Recently, Wang et al. ~\cite{wang2023prolificdreamer} proposed a variational score distillation (VSD) loss that is friendly to standard CFG guidance strength and better resolves unnatural textures. Instead of seeking a single data point, this approach regards the solution corresponding to a text prompt as a random variable. Specifically, VSD optimizes a distribution  $q^{\mu}(\bm{x}_0|y)$ of the possible 3D representations $\mu(\theta|y)$ corresponding to the text $y$, to be closely aligned with the distribution defined by the diffusion timestep $t=0$, $p(\bm{x}_0|y)$, in terms of KL divergence:
\begin{equation}
\mathcal{L}_\textnormal{VSD} = D_\textnormal{KL}(q^{\mu}(\bm{x}_0|y) || p(\bm{x}_0|y)).
\end{equation}
\cite{wang2023prolificdreamer} further shows that this objective can be optimized by matching the score of noisy real images and that of noisy rendered images at each time $t$, so the gradient of $\mathcal{L}_\textnormal{VSD}$ is
\begin{equation}
\scalebox{0.8}{$
\nabla_{\theta}\mathcal{L}_\textnormal{VSD}(\phi, g(\theta))=\mathbb{E}_{t, \bm{\epsilon}}\Big[\omega(t)(\bm{\epsilon}_{\phi}(\bm{x}_{t};y,t)-\bm{\epsilon}_\textnormal{lora}(\bm{x}_{t};y,t,c))\frac{\partial \bm{x}}{\partial \theta}\Big].
$}
\label{eq:vsd}
\end{equation}
Here, $\bm{\epsilon}_\textnormal{lora}$ estimates the score of the rendered images using a LoRA (Low-rank adaptation)~\cite{hu2021lora} model. The obtained variational distribution yields samples with high-fidelity textures. However, this loss is applied for texture enhancement and is helpless to the coarse geometry initially learned by SDS. Moreover, both the SDS and VSD attempt to distill from a fixed target 2D distribution which only assures per-view plausibility rather than a global 3D consistency. Consequently, they suffer from the same appearance and semantic shift issue that hampers the perceived 3D quality.
\section{DreamCraft3D++: Overview}

We propose a hierarchical pipeline for high-quality and efficient 3D content generation as illustrated in Figure~\ref{fig:pipeline}. Given a single image, our method first leverages state-of-the-art multi-view diffusion models to produce several orthogonal and consistent multi-view images. Then, we build a feed-forward sparse-view 3D reconstruction model (Sec. \ref{sec:lrm}) to efficiently infer the underlying textured meshes from those input images. For this stage, we prioritize multi-view consistency and global 3D structure, allowing for some compromise on detailed textures and geometry. Finally, we focus on jointly optimizing realistic and coherent texture as well as detailed geometry, with a training-free object-aware diffusion prior (Sec. \ref{sec:refiner}).

\section{MP-LRM: Multi-Plane large reconstruction model}
\label{sec:lrm}


\begin{figure*}[t]
\begin{center}
\includegraphics[width=0.9\linewidth]{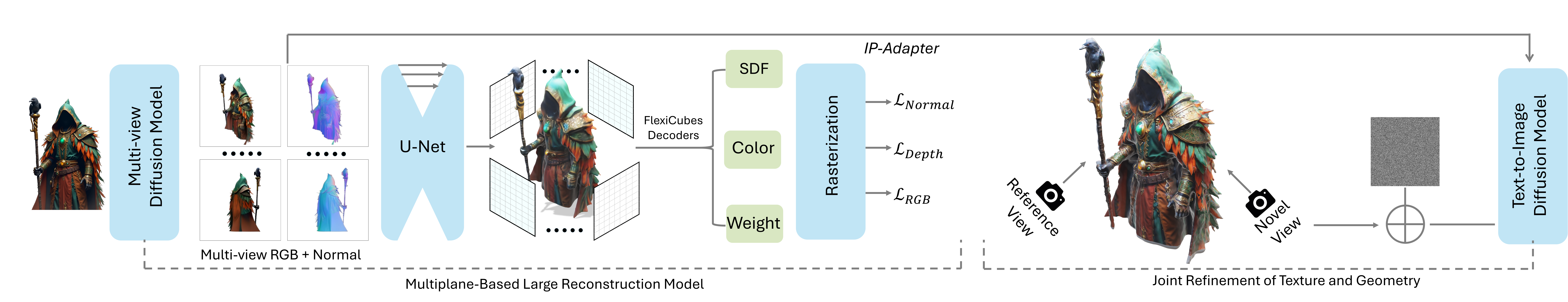}
\caption{DreamCraft3D++ pipeline. A single input image is processed by multi-view diffusion models to generate orthogonal, consistent views and normal maps. A feed-forward sparse-view 3D reconstruction model (Sec.~\ref{sec:lrm}) infers textured meshes from the multi-view images using a convolutional U-Net to map input to non-orthogonal planes, decoded into Flexicubes. Finally, a training-free object-aware diffusion prior enhances high-frequency geometry and texture details via score distillation (Sec.~\ref{sec:refiner}).}
\label{fig:pipeline}
\end{center}
\end{figure*}

As illustrated in Figure~\ref{fig:pipeline}, the Multi-Plane Large Reconstruction Model (MP-LRM) takes as input multi-view images and their corresponding normal maps with known camera poses. A convolutional U-Net is employed to map the input images and normal maps to a set of non-orthogonal multiple planes (Sec. \ref{sec:multiplane} and \ref{sec:learnableplane}). Subsequently, lightweight multi-layer perceptrons (MLPs) are utilized to decode the triplane features into signed distance field (SDF) values, texture colors, and Flexicubes parameters (Sec. \ref{sec:decoding}). Finally, these decoded values are used to obtain a textured mesh via the dual marching cubes algorithm. In the following subsections, we elaborate on the key components of MP-LRM (Sec. \ref{sec:lrm_loss}).

\subsection{Nonorthogonal multi-plane representation with U-Net based backbone} 
\label{sec:multiplane}
The core of our framework is a U-Net-based backbone $\mathcal{U}$ that predicts a nonorthogonal multi-plane representation from multi-view images. Figure~\ref{fig:pipeline} illustrates the network architecture. The input consists of multi-view images $N$ multi-view images \{${\mathbf{I}_i \in \mathbb{R}^{H \times W \times 3} | i = 1,2,...,N}$\} and their corresponding normal maps \{${\mathbf{N}_i \in \mathbb{R}^{H \times W \times 3} | i = 1,2,...,N}$\} with known poses. Here we adopt Zero123++~\cite{shi2023zero123++} to generate 6 fix-viewed images and the number of planes $N=6$. Following previous works~\cite{gslrm2024,xu2024grm}, we employ Plücker ray embedding to densely encode the camera poses. The RGB values and ray embeddings are concatenated into 12-channel feature maps, which are then fed through $\mathcal{U}$ to predict a set of nonorthogonal planes \{${\mathbf{\Pi}_i \in \mathbb{R}^{H \times W \times C} | i = 1,2,...,N}$\}. The term ``nonorthogonal'' refers to coordinate systems where the axes are not perpendicular to each other, unlike the axis-aligned triplanes used in other methods. We opt for nonorthogonal planes instead of orthogonal axis-aligned planes because we believe that learning the mapping from 2D posed images to orthogonal plane features is inefficient. If the network capacity is insufficient, the mapping result is prone to blurriness. Considering that multi-view diffusion models always generate fixed-view images, we allow the network to focus on learning pixel-aligned features without the need to unpose them in space. 

We employ a convolutional U-Net $\mathcal{U}$ to learn the mapping from 2D posed images to non-orthogonal plane features. Compared to transformer-based methods~\cite{gslrm2024,hong2023lrm,xu2024instantmesh,PF-LRM}, our U-shape design has a larger bandwidth for preserving input information, resulting in highly detailed triplane features and, ultimately, elaborate textured meshes. Moreover, the convolutional network fully utilizes the geometric prior of the spatial correspondence between triplanes and the input six orthographic images, which greatly accelerates convergence and stabilizes training. Our model $\mathcal{M}$ can achieve reasonable reconstruction results at a very early stage of training (around 20 minutes of training from scratch).

\subsection{Learnable plane embeddings}
\label{sec:learnableplane}

To address the lack of information in top and bottom views due to the limited elevation range of generated multi-view images, which often results in meshes with holes or artifacts, we introduce two additional learnable plane embeddings. These embeddings, denoted as $\mathbf{E}_{\text{top}}$ and $\mathbf{E}_{\text{bottom}}$, are strategically placed at the top and bottom, respectively. They are processed through a U-Net-based backbone $\mathcal{U}_E$. Specifically, these learnable plane embeddings and input images are processed separately by the corresponding U-Net model's down-sampling and up-sampling paths, while the intermediate feature maps from both U-Nets are concatenated and passed through the self-attention module. Our experiments demonstrate that this approach effectively mitigates holes, enhancing the quality of the reconstructed meshes.

\subsection{Decoding planes to Flexicubes} 
\label{sec:decoding}
Previous generic 3D generation methods predominantly employ NeRF~\cite{mildenhall2021nerf} or Gaussian splatting~\cite{kerbl2023gaussian} as the geometry representation. These methods rely on additional procedures, such as Marching Cubes (MC), to extract the iso-surface, leading to topological ambiguities and challenges in representing high-fidelity geometric details. In this work, we utilize Flexicubes~\cite{shen2023flexible} as our geometry representation. Flexicubes allow for mesh extraction from grid features using dual marching cubes during training. The features include signed distance function (SDF) values, deformation, and weights. Texture is obtained by querying the color at the surface. Flexicubes enable the training of our reconstruction model to produce textured meshes as the final output in an end-to-end manner. Given the surface vertices, we project them onto each nonorthogonal plane and query the feature using bilinear sampling. The features from all planes are aggregated through channel-wise concatenation.

\subsection{Loss Function}
\label{sec:lrm_loss}
\noindent\textbf{RGB loss.}
During training, we render the images at the $K$ supervision views, and minimize the image reconstruction loss.
Let $\{I_{i}^{gt} | i = 1, 2,..., K\}$ be the set of groundtruth views, and $\{\hat{I}_{i}\}$ be the rendered images, our loss function 
is a combination of MSE (Mean Squared Error) loss and Perceptual loss:
\begin{equation}
\begin{aligned}
    \mathcal{L}_{\mathrm{rgb}} &= \mathrm{\lambda_{mse}} \cdot \frac{1}{K} \sum_{i=1}^K \left\|\hat{I}_{i},I_{i}^{gt}\right\|_2^2 \\
    &+ \mathrm{\lambda_{lpips}} \cdot \frac{1}{K} \sum_{i=1}^K \mathcal{L}_{\mathrm{lpips}}\left(\hat{I}_{i},I^{gt}_{i}\right)
\end{aligned}
\end{equation}

where $\mathrm{\lambda_{lpips}}$ are the weight of the Perceptual loss. During training, we set $\mathrm{\lambda_{lpips}} = 2.0$.

\noindent\textbf{Mask loss.} 
During training, we also render the masks at the $K$ supervision views, and minimize the mask loss.
Let $\{M_{i}^{gt} | i = 1, 2,..., K\}$ be the set of groundtruth views, and $\{\hat{M}_{i}\}$ be the rendered masks, and we adopt Binary Cross Entropy (BCE) loss as mask loss:
\begin{equation}
    \mathcal{L}_{\mathrm{mask}} = \mathrm{\lambda_{mask}} \cdot \frac{1}{K} \sum_{i=1}^K \mathcal{L}_{\mathrm{BCE}}\left(\hat{M}_{i},M^{gt}_{i}\right)
\end{equation}

where $\mathrm{\lambda_{mask}}$ are the weight of the mask loss. During training, we set $\mathrm{\lambda_{mask}} = 0.1$.

\noindent\textbf{Depth and normal loss.} 
In addition, akin to Nerdi~\cite{deng2023nerdi}, we fully exploit the geometry prior inferred from the reference image, and enforce the consistency with the depth and normal map computed for the reference view. The corresponding  depth and normal loss are respectively computed as:

\begin{align}
    \mathcal{L}_{\mathrm{depth}} &= \mathrm{\lambda_{depth}} \cdot \frac{1}{K} \sum_{i=1}^K M_{i}^{gt} \otimes \left\|\hat{D}_{i},D_{i}^{gt}\right\|_1 \\
    \mathcal{L}_{\mathrm{normal}} &= \mathrm{\lambda_{normal}} \cdot \frac{1}{K} \sum_{i=1}^K M_{i}^{gt} \otimes \left(1 - \hat{N}_{i} \cdot N_{i}^{gt}\right)
\end{align}




\noindent The overall loss function is:
\begin{equation}
    \mathcal{L}_{\mathrm{total}} = \mathcal{L}_{\mathrm{rgb}} + \mathcal{L}_{\mathrm{mask}} + \mathcal{L}_{\mathrm{depth}} + \mathcal{L}_{\mathrm{normal}}
\end{equation}

\section{Accelerated Joint Refinement of Texture and Geometry}
\label{sec:refiner}

\subsection{Training-free object-aware diffusion prior}
The textured meshes generated by our MP-LRM, while highly consistent with the multi-view input images, still suffer from detail distortion due to two main factors: 1) limited network generalization caused by insufficient training data; and 2) inconsistent and blurry multi-view renderings produced by the multi-view diffusion models. To address these issues, we introduce a refinement algorithm that simultaneously enhances both texture and geometry.

Previous methods refine 3D details by optimizing the SDS loss, However, these pretrained 2D diffusion priors lack object awareness, resulting in inconsistent optimization during 3D reconstruction. DreamCraft3D tackles this problem by fine-tuning a pretrained diffusion model using multi-view image renderings through DreamBooth~\cite{ruiz2023dreambooth}. This approach allows the model to form a concept of the scene's surrounding views and promotes consistent texture generation. Nevertheless, training DreamBooth is computationally inefficient due to its reliance on extensive iterative processes. Its susceptibility to overfitting and the need for careful parameter tuning further increase resource demands and training time.

To overcome this limitation, we are the first to propose the use of IP-Adapter~\cite{ye2023ip}, a training-free, lightweight image prompt adaptation method that employs a decoupled cross-attention strategy for existing text-to-image diffusion models. Unlike DreamBooth, Ip-adapter does not require training, making it significantly more efficient and suitable for our refinement algorithm. By leveraging Ip-adapter, our method can efficiently refine the textured meshes, enhancing both texture and geometry while maintaining consistency with the multi-view input images. Notably, utilizing IP-Adapter allows our method to refine textured meshes four times faster than DreamCraft3D using DreamBooth, reducing the optimization time from 40 minutes to just 10 minutes.

To mitigate ambiguity and texture multi-head problems that arise when conditioning IP-Adapter on only the source image, we condition the model on both the source image ${\mathbf{I}_{source}}$ and the generated multi-view renderings ${\mathbf{I}_i}, i = 1,2,...,N$. Similar to DreamCraft3D, we employ an off-the-shelf upsampler~\cite{yu2024scaling} to augment multi-view renderings before conditioning. We obtain image embeddings from the IP-Adapter for all view images. During training, the IP-Adapter image embedding is selected based on the location of the randomly sampled camera, called \textit{view-dependent image prompting}. We use a weighted combination of the image embeddings of different views depending on the value of the azimuth angle $\theta_{cam}$. This approach ensures that the most relevant image embedding is utilized based on the camera's position, enhancing the consistency and quality of the generated 3D reconstructions.

\subsection{Joint texture-geometry refinement}

\noindent\textbf{Camera and light augmentations.} We follow Magic3D~\cite{lin2023magic3d} to add random augmentations to the camera and light sampling for rendering the shaded images. Differently, we sample the point light location such that the angular distance from the random camera center location (w.r.t. the origin) is sampled with a random point light distance $r_\textnormal{cam}$ in [7.5, 10], and we freeze the material augmentation unlike Dreamfusion and Magic3D, as we found it is bad for training convergence. During training, we fix the Field-of-View angle to $40^\circ$, and sample elevation angle $\phi_{cam}$ from $\mathcal{U}(-10,45)$ and azimuth angle $\theta_{cam}$ from $\mathcal{U}(-180,180)$, and distance from the origin in [1, 1.2].

\noindent\textbf{Iterative RGB and Normal rendering.} 
DreamCraft3D alternately renders normal maps $\hat{N}$ and RGB images $\hat{I}$ as the input for diffusion guidance $\hat{I}_{g}$ to enhance texture and geometry disentanglement. For convenience, we ignore the view subscript. While this strategy leads to finer geometry, it can result in geometry-texture misalignment due to the absence of RGB cues in normal maps, causing divergent optimization directions for geometry and texture. To address this issue, we propose a new strategy that blends normal maps and RGB images using a random weight $\alpha \in [0, 1]$, instead of solely relying on normal maps for guidance input: 

\begin{equation}
\hat{I}_{g} = 
\begin{cases}
\hat{I} & \text{if } r < 0.5 \\
\alpha \cdot \hat{I} + (1 - \alpha) \cdot {\hat{N}} & \text{if } r \geq 0.5
\end{cases}
\end{equation}

where $r \sim \text{Uniform}(0, 1)$ and $\alpha \sim \text{Uniform}(0, 0.5)$ is a random variable used to weight the RGB and normal components of the output when $r \geq 0.5$. This blending approach incorporates both geometric information from the normal maps and color cues from the RGB images, promoting a more coherent optimization of geometry and texture.

\subsection{Loss functions}
We supervise the refinement by two parts: pixel-level loss under the reference view and the diffusion distillation loss at random views. For the diffusion distillation loss, since we adopt a training-free customized diffusion prior, the standard VSD loss in \cite{wang2023prolificdreamer} is used instead of BSD in DreamCraft3D: 
\begin{equation}
\small
\nabla_{\theta}\mathcal{L}_\textnormal{VSD}(\theta)=\mathbb{E}_{t, \bm{\epsilon}, c}[\omega(t)(\bm{\epsilon}_\textnormal{ipa}(\hat{I}_{t};y^c,t)-\bm{\epsilon}_\textnormal{lora}(\hat{I}_{t};t,c,y))\frac{\partial \hat{I}}{\partial \theta}],
\label{eq:vsd}
\end{equation}

\noindent where $\hat{I}_t = \alpha_t \hat{I} + \sigma_t\bm{\epsilon}$. $\bm{\epsilon}_\textnormal{ipa}(\hat{I}_{t};y^c,t)$ is a pretrained diffusion model adapted to multi-view image prompts using IP-Adapter. $\bm{\epsilon}_\textnormal{lora}$ is parameterized by a LoRA (Low-rank adaptation~\cite{hu2021lora}) of $\bm{\epsilon}_\textnormal{ipa}(\hat{I}_{t};y^c,t)$, conditioned on additional camera parameter c. 

The pixel-level reconstruction loss at reference view $\mathcal{L}_{recon}$ is the combination of RGB MSE loss and LPIPS loss, as well as consistency loss~\cite{munkberg2022extracting}:
\begin{equation}
\small
    \mathcal{L}_{total} = \lambda_{rgb}\mathcal{L}_{mse} + \lambda_{lpips}\mathcal{L}_{lpips} + \lambda_{consistency}\mathcal{L}_{consistency} 
\end{equation}
\noindent The total loss is:
\begin{equation}
        \mathcal{L}_{total} = \mathcal{L}_{recon} + \lambda_{vsd}\mathcal{L}_{vsd},
\end{equation}

\noindent where $\lambda_{rgb} = \lambda_{lpips} = 10000$, $\lambda_{consistency} = 100$, $\lambda_{vsd} = 0.1$. 
\section{Experiments}

\subsection{Implementation Details}
\noindent\textbf{Dataset.} We train our model on selected partitation of Objaverse dataset~\cite{deitke2024objaverse}, which contains 240k 3D models. We use Blender to render ground-truth $512\times512$ images, depths and normals for an object. We normalize the shape to the box [-0.5, 0.5] in world space and render 100 random views and 6 fixed views aligned with Zero123++ V2 setting, lit with randomly selected environment maps.

For Evaluation, we adopt Google Scanned Objects (GSO) dataset~\cite{downs2022google}, which includes a wide variety of high-quality scanned household items, to evaluate the performance of our method and other baselines. Following InstantMesh~\cite{xu2024instantmesh}, we randomly choose 300 shapes and render 21 images of each object in an orbiting trajectory with uniform azimuths and varying elevations in ${30^\circ, 0^\circ, 30^\circ}$. Besides, we establish a test benchmark that includes 300 images, which is a mix of real pictures and those produced by Stable Diffusion~\cite{rombach2021highresolution} and Deep Floyd. Each image in this benchmark comes with an alpha mask for the foreground, a predicted depth map, and a text prompt. For real images, the text prompts are sourced from an image caption model. We intend to make this test benchmark accessible to the public.

\noindent\textbf{Architectural details.} Our model MP-LRM contains 600M parameters, with two U-Nets for the multi-view images and learnable plane embeddings. The U-Nets have [64, 128, 128, 256, 256, 512, 512] channels and attention blocks at resolutions [64, 32, 16]. We generate 6-view 256$\times$256$\times$36 images and normal maps using Zero123++~\cite{shi2023zero123++}. The two learnable 256$\times$256$\times$32 plane embeddings are placed at the bottom and top. The plane decoder is of 5 layers with hidden dimensions 64. The Flexicubes grid size is 96 and the rendering resolution is 512 for both training and inference.

We employ DreamShaper~\cite{dreamshaper2023} as the pretrained text-to-image diffusion model for guidance. For image prompt adaptation, we initially utilize SUPIR~\cite{yu2024scaling} to enhance the resolution of the multi-view images generated by Zero123++ and then apply IP-Adapter\cite{ye2023ip} to adapt the diffusion model to those upscaled images and the original input image. The IP-Adapter scale is configured to 0.8.

\noindent\textbf{Training details.} 
We train MP-LRM using 64 NVIDIA A100 (80G) GPUs with a batch size of 256 for 100 epochs, which takes approximately 5 days to complete. For each training sample, we use 6 fixed views aligned with Zero123++ as input images. To supervise the shape reconstruction, we utilize a total of 4 views: 3 randomly selected from a set of 100 views and 1 randomly chosen from the input views. During training, we use random background color augmentation. We optimize the model using the AdamW optimizer with a learning rate of $4 \times 10^{-4}$ and a cosine learning rate schedule.

For the refinement, we build our system based on the foundation of threestudio~\cite{threestudio2023}. We improve the Flexicubes grid size from 96 to 192 for detail sculpting. We optimize each case for 2000 iterations on one NVIDIA A100 (80G) GPU with batch size of 1. We optimize the model using the AdamW optimizer with separate learning rates for different modules: $\eta_{\mathbf{\epsilon}_\text{lora}}=1e^{-4}$, $\eta_{\mathcal{G}_{e}}=1e^{-2}$, $\eta_{\mathcal{F}}=1e^{-3}$. $\mathbf{\epsilon}_\text{lora}$, $\mathcal{G}_{e}$, $\mathcal{F}$ represent the LoRA layer of the guidance, geometry encoding network, and Flexicubes parameters. During training, the diffusion timestep $t$ is sampled from range [0.02, 0.4]. 

\subsection{Comparisons with the State of the Arts}

\noindent\textbf{Baselines.} We compare our technique against six baselines: four feed-forward methods (TripoSR~\cite{Tochilkin2024TripoSRF3}, LGM~\cite{Tang2024LGMLM}, CRM~\cite{Wang2024CRMSI}, InstantMesh~\cite{xu2024instantmesh}) and one optimization-based method (DreamCraft3D~\cite{sun2023dreamcraft3d}). TripoSR is the best-performing open-source LRM for single-view reconstruction. LGM and CRM are UNet-based models that reconstruct Gaussians and 3D meshes from multi-view images, respectively. InstantMesh is a Transformer-based LRM using Flexicubes. DreamCraft3D optimizes 3D reconstruction from a single image.

\newcommand{\mrka}[1]{{\colorbox{red!30}{#1}}}
\newcommand{\mrkb}[1]{{\colorbox{red!20}{#1}}}
\newcommand{\mrkc}[1]{{\colorbox{red!10}{#1}}}
\begin{table}[t]
  \small
  \renewcommand{\tabcolsep}{1.5mm}
  \caption{Quantitative results on Google Scanned Objects (GSO) orbiting views.}
  \centering
  \begin{tabular}{@{}l|c|c|c|c|c}
    \toprule
     Method & PSNR $\uparrow$ & SSIM $\uparrow$ & LPIPS $\downarrow$ & CD $\downarrow$ & FS $\uparrow$ \\
    \midrule
    TripoSR & 14.152 & 0.829 & 0.213 & 0.168 & 0.730 \\
    LGM & 13.814 & 0.821 & 0.218 & 0.191 & 0.642 \\
    CRM & 16.401 & 0.838 & 0.201 & 0.166 & 0.754 \\
    InstantMesh & \mrkc{16.697} & \mrkc{0.850} & \mrkc{0.157} & \mrkc{0.140} & \mrkb{0.804} \\
    DreamCraft3D & 16.302 & 0.835 & 0.230 & 0.177 & 0.656 \\
    Ours (Coarse) & \mrkb{17.729} & \mrkb{0.843} & \mrkb{0.151} & \mrka{0.131} & \mrka{0.821}  \\
    Ours & \mrka{20.251} & \mrka{0.862} & \mrka{0.132} & \mrkb{0.136} & \mrkc{0.795} \\
    \bottomrule
  \end{tabular}
  \label{tab:gso}
\end{table}

\begin{figure*}[!h]
\begin{center}
\includegraphics[width=0.9\textwidth]{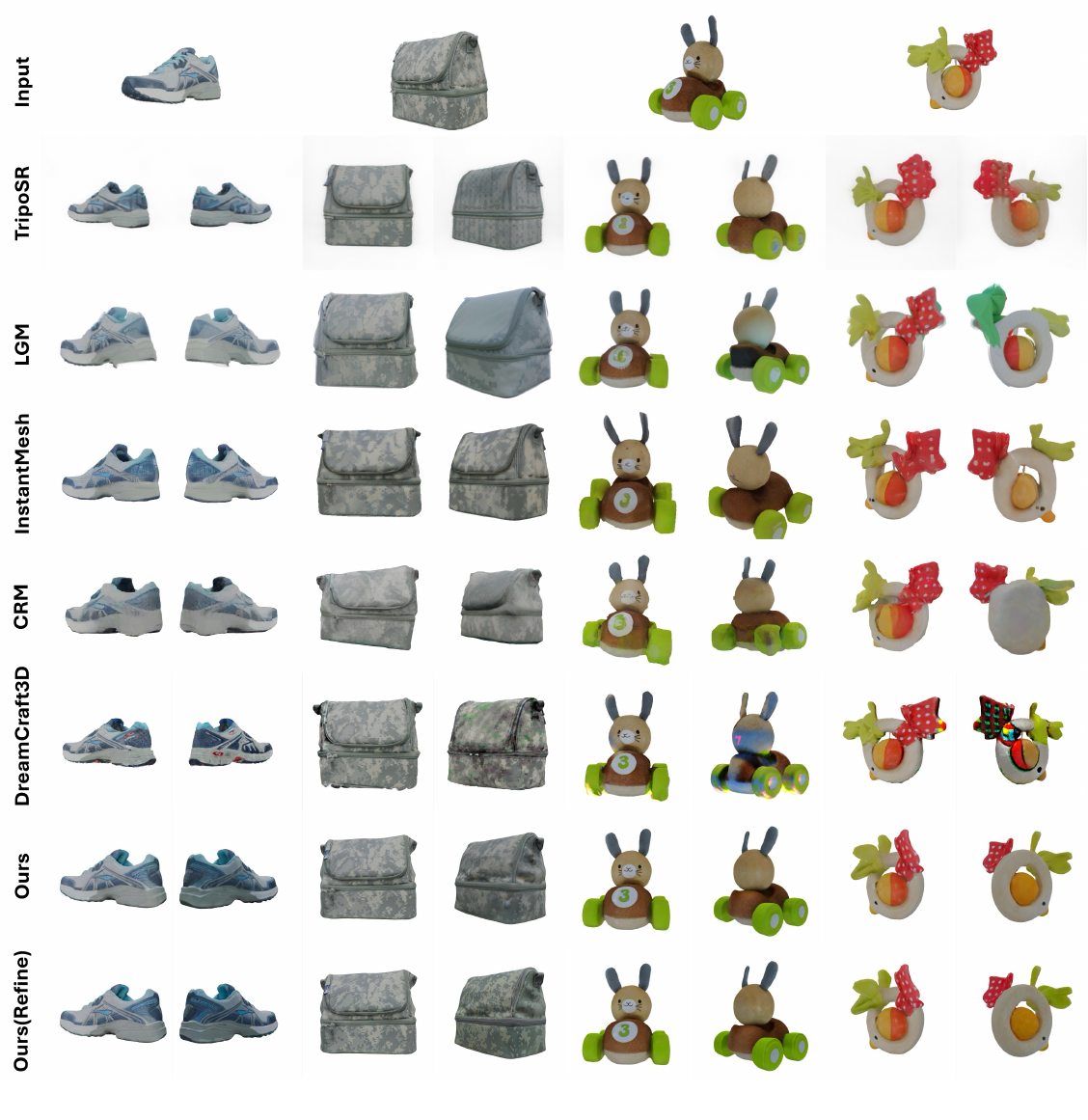}
\caption{Qualitative comparison with baselines on the GSO dataset.}
\label{fig:gso}
\end{center}
\end{figure*}

\begin{figure*}[!h]
\begin{center}
\includegraphics[width=0.9\textwidth]{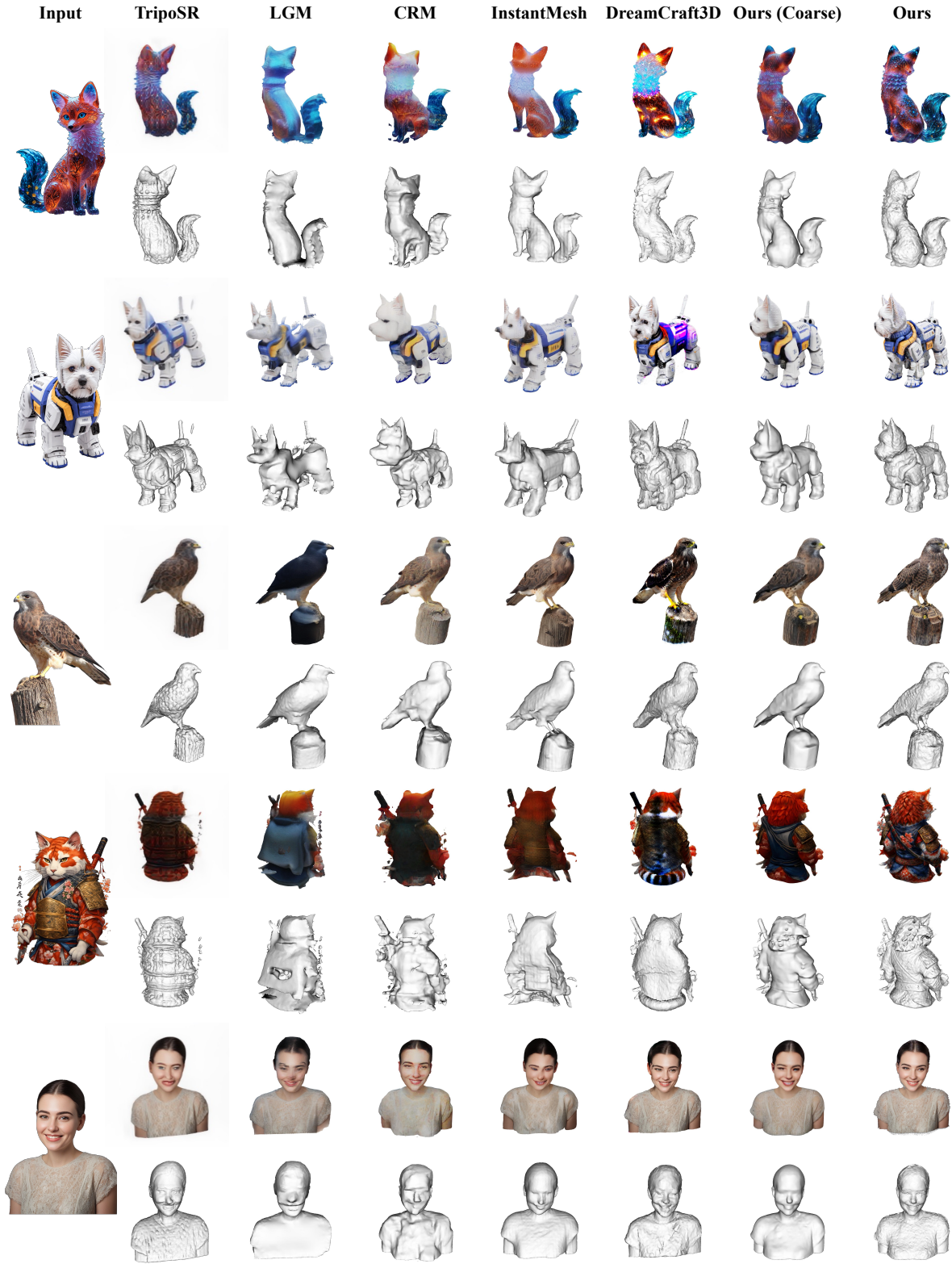}
\caption{Qualitative comparison with baselines on the Internet images.}
\label{fig:internet}
\end{center}
\end{figure*}

\noindent\textbf{Quantitative comparison.}
We report the quantitative results on GSO~\cite{downs2022google} dataset in Table ~\ref{tab:gso}. For each metric, we highlight the top three results
among all methods, and a deeper color indicates a better
result. We use four evaluation metrics: PSNR, SSIM, LPIPS, Chamfer Distance (CD) and F-Score (with a threshold of 0.05). The first four metrics are for novel view synthesis and the last two are for geometry reconstruction quality. 

From the 2D novel view synthesis metrics, we can observe that DreamCraft3D++, including our coarse model, outperforms the baselines on PSNR, SSIM and LPIPS significantly, indicating that its generation results have the best perceptually viewing quality.

As for the 3D geometric metrics, DreamCraft3D++ outperforms the baselines on both CD and FS, indicates more reliable generated shapes. Note that the result after refinement has slightly worse geometry metric performance because the objects optimize more surface detail guided by the pretrained diffusion model, which may affect chamfer distance.

\noindent\textbf{Qualitative comparison.} To validate the effectiveness of our method, we qualitatively compare our results with the baselines: TripoSR, LGM, CRM, InstantMesh, and DreamCraft3D. For all baselines, we use their official code and checkpoints. We first visualize images from the GSO dataset, as shown in Figure~\ref{fig:gso}. The figure demonstrates that our MP-LRM generates higher quality results compared to all other baselines. This can be attributed to the efficiency of our method, which fully utilizes the spatial alignment of input posed images and outputs a multiplane-based representation. Furthermore, the refinement stage restores texture details, further improving fidelity.

We also visualize the meshes generated from a single input image using these baselines on a set of more complex internet images, as shown in Figure~\ref{fig:internet}. Our improvement in 3D generation quality is even more evident in these complex cases. Compared to the one-stage methods, our MP-LRM produces smoother and more reasonable geometry with better detail. For instance, in the case of the fox and cat (second and eighth rows), LGM, CRM, and InstantMesh generate incomplete meshes, while our coarse results are complete and smooth. More importantly, the significance of our proposed refinement stage is clear. By leveraging a powerful 2D pre-trained diffusion model, it enhances texture and geometric details while maintaining a unified style consistent with the input image through multi-view conditioned object-awareness.

DreamCraft3D generates sharper texture details compared to the one-stage methods; however, it still falls short of our approach. We posit that this is due to the sensitivity of DreamBooth training to hyperparameters and training steps, which can lead to over-saturated textures. Furthermore, the iterative nature of DreamBooth and 3D refinement in DreamCraft3D may exacerbate this issue, causing the accumulation of artifacts over multiple iterations.

\subsection{Ablation study}
\noindent\textbf{The learnable plane embeddings.} In our paper, in order to address the lack of information in the top and bottom views, we introduce two additional learnable plane embeddings placed at the top and bottom of the volume, respectively. Table~\ref{tab:embedding} demonstrates that this design effectively mitigates this issue by incorporating the additional embeddings, thereby reducing holes and enhancing the overall quality of the reconstructed meshes.

\begin{figure}[t]
\begin{center}
\includegraphics[width=1.0\linewidth]{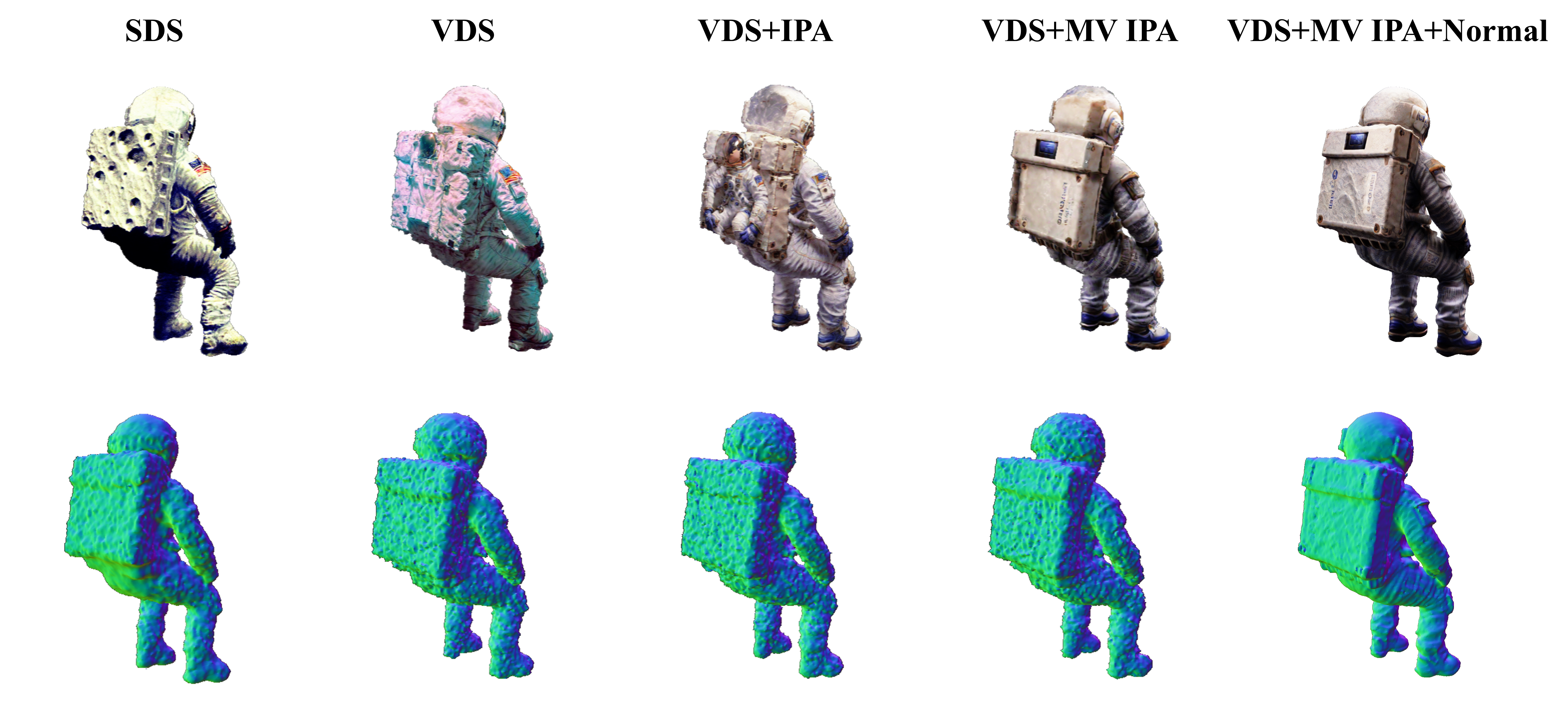}
\caption{Ablation study on the strategies of the refinement.}
\label{fig:sds}
\end{center}
\end{figure}

\begin{figure}[t]
\begin{center}
\includegraphics[width=1.0\linewidth]{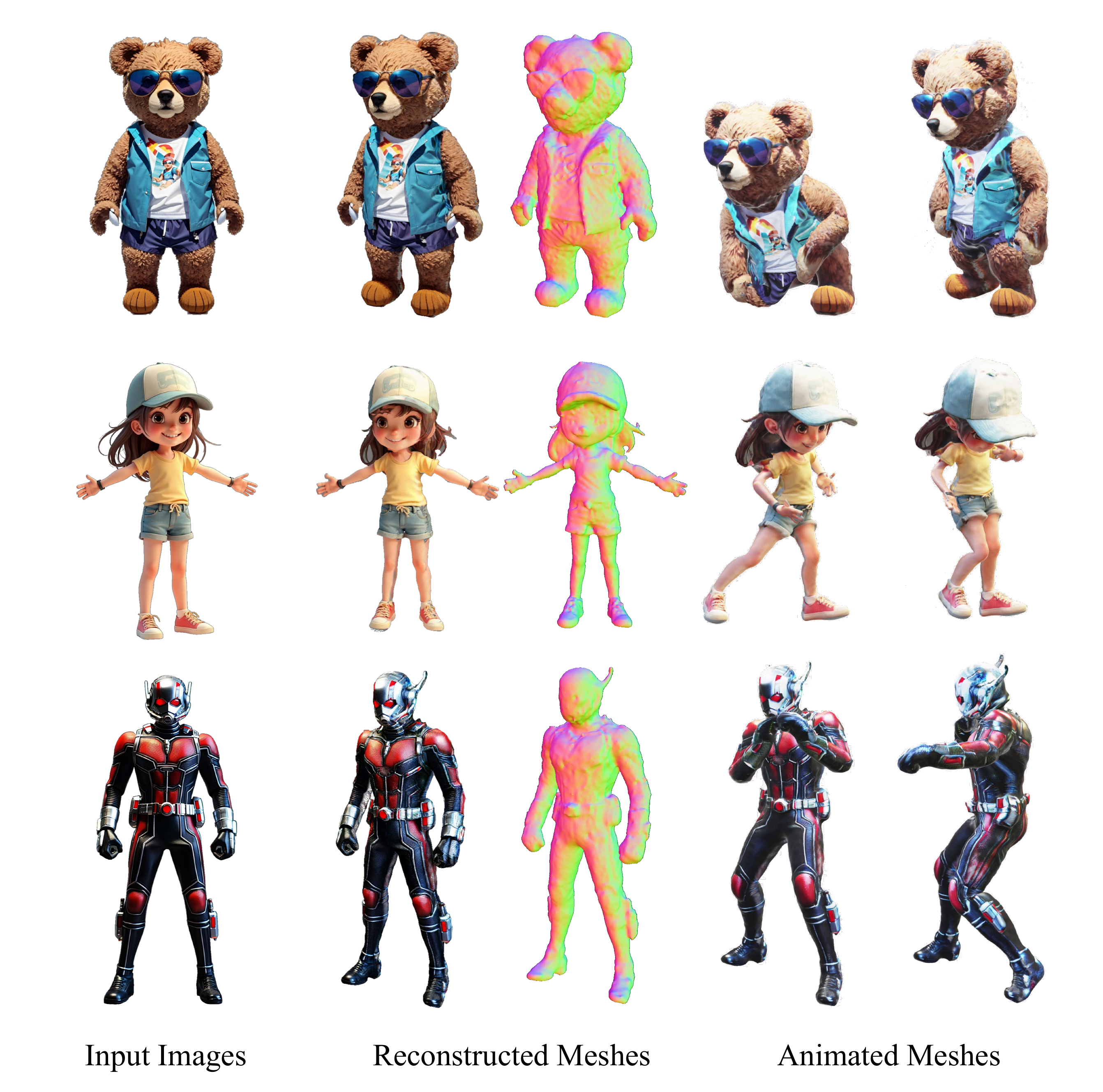}
\caption{We export the high-quality textured meshes, enabling seamless integration into downstream applications such as 3D rigged animation.}
\label{fig:animation}
\end{center}
\end{figure}

\begin{table}[t]
  \small
  \renewcommand{\tabcolsep}{1.5mm}
  \caption{Ablation study on the learnable plane embeddings}
  \centering
  \begin{tabular}{@{}l|c|c|c|c|c}
    \toprule
     Method & PSNR $\uparrow$ & SSIM $\uparrow$ & LPIPS $\downarrow$ & CD $\downarrow$ & FS $\uparrow$ \\
    \midrule
    w/o $\mathbf{E}$ & 17.119 & 0.838 & 0.174 & 0.152 & 0.754  \\
    Ours & 17.729 & 0.843 & 0.151 & 0.131 & 0.821 \\
    \bottomrule
  \end{tabular}
  \label{tab:embedding}
\end{table}

\noindent\textbf{The strategies of the refinement.} Figure~\ref{fig:sds} presents an ablation study comparing five texture optimization techniques: (1) Score Distillation Sampling (SDS), (2) Variational Score Distillation (VSD), (3) VSD with a single-view image-conditioned IP-Adapter, (4) VSD with a multi-view image-conditioned IP-Adapter, and (5) VSD with a multi-view image-conditioned IP-Adapter and iterative normal-image rendering.
SDS generates novel-view textures that appear overly smooth and saturated. While VSD using standard stable diffusion produces more realistic textures, it suffers from notable inconsistencies, such as the astronaut's face appearing at the back view. Incorporating a single-view image-conditioned IP-Adapter introduces object awareness but still results in multi-face texture issues.
Our proposed multi-view image-conditioned approach achieves a balance between realism and consistency, although the resulting geometry appears noisy. To address this, we employ iterative RGB and normal rendering, which enhances the disentangled optimization of texture and geometry and refines the surface geometry.

\subsection{Applications}

\noindent\textbf{3D rigged animation.} As shown in Figure~\ref{fig:animation}, we export the high-quality textured meshes for downstream applications, including rigged animation. By utilizing software such as Mixamo~\cite{Mixamo}, we can efficiently employ these meshes for rigging and animation tasks, thereby streamlining the workflow and improving the overall production process.




\section{Discussions and Conclusions}
\noindent\textbf{Limitations.} Though DreamCraft3D++ demonstrates impressive capabilities in high-quality 3D generation, it is not without limitations. A significant drawback lies in the quality of multi-view images produced by Zero123++, which are directly utilized for 3D reconstruction via MP-LRM and IP-Adapter prompt conditioning. Zero123++ struggles to generate satisfactory multi-view images when presented with complex inputs or significant elevation angles. Additionally, DreamCraft3D++ outputs 3D objects with baked illumination, rendering them unsuitable for graphics pipelines that require controlled lighting conditions.

\noindent\textbf{Future work.} For future work, one promising direction involves the exploration of enhanced multi-view diffusion models that can deliver higher quality outputs and accommodate a broader range of elevation angles in input images. Furthermore, integrating physically-based rendering (PBR) materials into the 3D generation process could yield significant improvements. Finally, expanding the scope of 3D generation from individual objects to entire scenes is essential, particularly by supporting flexible input formats such as captured video sequences or multiple unposed images.

\noindent \textbf{Conclusions.} In this work, we present DreamCraft3D++, a framework for efficient high-quality generation of complex 3D assets. Building on the multi-stage process of DreamCraft3D, we replace the time-consuming geometry sculpting optimization with a feed-forward, multi-plane reconstruction model, achieving a 1000x speedup. For texture refinement, our training-free IP-Adapter module utilizes enhanced multi-view images to improve texture and geometry consistency, providing a solution that is four times faster than DreamCraft3D's DreamBooth fine-tuning. Our approach generates intricate 3D assets with realistic 360° textures, significantly outperforming current state-of-the-art image-to-3D methods in quality and speed.

\bibliographystyle{IEEEtran}
\bibliography{reference}

\end{document}